\documentclass[sigconf]{acmart}

\usepackage{epsfig}
\usepackage{graphicx}
\usepackage{amsmath}
\usepackage{caption}
\usepackage{subcaption}
\usepackage{xspace}

\makeatletter
\DeclareRobustCommand\onedot{\futurelet\@let@token\@onedot}
\def\@onedot{\ifx\@let@token.\else.\null\fi\xspace}

\def\eg{\emph{e.g}\onedot} 
\def\ie{\emph{i.e}\onedot} 
 
\def\etc{\emph{etc}\onedot} \def\vs{\emph{vs}\onedot}
 
\def\etal{\emph{et al}\onedot}
\makeatother

\AtBeginDocument{%
  \providecommand\BibTeX{{%
    \normalfont B\kern-0.5em{\scshape i\kern-0.25em b}\kern-0.8em\TeX}}}

\copyrightyear{2020}
\acmYear{2020}
\setcopyright{acmcopyright}\acmConference[SUMAC'20]{2nd Workshop on Structuring and Understanding of Multimedia heritAge Contents}{October 12, 2020}{Seattle, WA, USA}
\acmBooktitle{2nd Workshop on Structuring and Understanding of Multimedia heritAge Contents (SUMAC'20), October 12, 2020, Seattle, WA, USA}
\acmPrice{15.00}
\acmDOI{10.1145/3423323.3423407}
\acmISBN{978-1-4503-8155-0/20/10}

\acmSubmissionID{7}

\setcopyright{none}
\begin{document}
\fancyhead{}

%
\title{PP-LinkNet: Improving Semantic Segmentation of High Resolution Satellite Imagery with Multi-stage Training}

%

\author{An Tran}
\affiliation{%
  \institution{Grabtaxi Holdings}
  \streetaddress{Institution1 address}
}
\email{an.tran@grab.com}

\author{Ali Zonoozi}
\affiliation{%
  \institution{Grabtaxi Holdings}
  \streetaddress{Institution2 address}
}
\email{ali.zonoozi@grab.com	}

\author{Jagannadan Varadarajan}
\affiliation{%
	\institution{Grabtaxi Holdings}
	\streetaddress{Institution2 address}
}
\email{jagan.varadarajan@grab.com}

\author{Hannes Kruppa}
\affiliation{%
	\institution{Grabtaxi Holdings}
	\streetaddress{Institution2 address}
}
\email{hannes.kruppa@grab.com }

%

%
\begin{abstract}
Road network and building footprint extraction is essential for many applications such as updating maps, traffic regulations, city planning, ride-hailing, disaster response \textit{etc}. Mapping road networks is currently both expensive and labor-intensive. Recently,  improvements in image segmentation through the application of deep neural networks has shown promising results in extracting road segments from large scale, high resolution satellite imagery.  However, significant challenges remain due to lack of enough labeled training data needed to build models for industry grade applications. In this paper, we propose a two-stage transfer learning technique to improve robustness of semantic segmentation for satellite images that leverages noisy pseudo ground truth masks obtained automatically (without human labor) from crowd-sourced OpenStreetMap (OSM) data. We further propose Pyramid Pooling-LinkNet (PP-LinkNet),  an improved deep neural network for segmentation that uses focal loss, poly learning rate, and context module. We demonstrate the strengths of our approach through evaluations done on three popular datasets over two tasks, namely, road extraction and building foot-print detection. Specifically, we obtain 78.19\% meanIoU on SpaceNet building footprint dataset, 67.03\% and 77.11\% on the road topology metric  on SpaceNet and DeepGlobe road extraction dataset, respectively.
\end{abstract}

%
%


%

%
\begin{teaserfigure}
  \includegraphics[width=\textwidth]{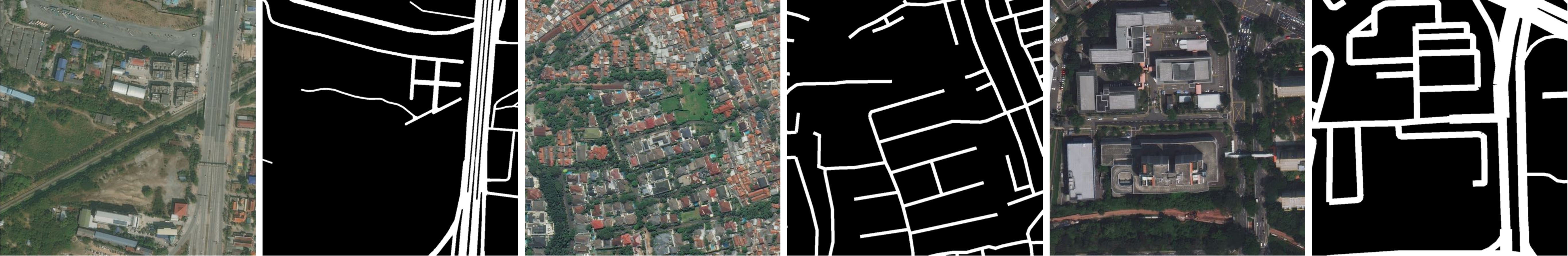}
  \label{fig:main examples}
\end{teaserfigure}

%
\maketitle

\section{Introduction}
Maps are common heritage for humanity. Creating and updating maps is an important task with many potential applications in disaster recovery, geospatial search, urban planning, ride-hailing industry, etc \cite{ijgi7030084}. Today, map features such as roads, building footprints, and points of interest are primarily created through manual techniques. We address the problem of extracting road networks and building footprints automatically from satellite imagery in this paper.

Maps of roads and buildings are expensive to build and maintain. Although modern cartography involves using satellite and aerial imagery along with GPS traces, using these data to update maps involves having humans analyze the data in a time consuming process, and thus maps of rapidly growing cities (where infrastructure is constantly under construction) are still often inaccurate and incomplete outside the urban core. Current approaches to map inference via satellite images~\cite{VanEtten2018,Demir2018} involve extensive annotations of images that is both tedious and time consuming. 

Currently, we have two large public datasets containing satellite images of road networks with annotations: i) SpaceNet \cite{VanEtten2018} and DeepGlobe \cite{Demir2018}. SpaceNet challenge 3 road extraction \cite{VanEtten2018} only provides 2779 training satellite images of Las Vegas, Paris, Shanghai, Khartoum with image size of $1300\times1300$ with their road center-line annotations. ii) DeepGlobe \cite{Demir2018} has a total of 8570 images with 6226 training, 1243 validation and 1101 testing images with pixel-based annotations, where all pixels belonging to the road are labeled, instead of labeling only center-line. Although seemingly large, these datasets are still not enough to train a robust model for analyzing satellite imagery on a global scale due to challenges posed by spatial variations -- roads differ in their appearance due to regional terrain, urban vs rural divide, state of economy, e.g., developed \vs developing countries, see Figure~\ref{fig: rasterization output samples})
and temporal variations resulting in images captured during different seasons with varying lighting conditions, cloud cover, \etc. In this paper, we develop an efficient and effective technique to fine-tune semantic segmentation models robustly on different variances of satellite images. The centered idea is that we leverage OpenStreetMap(OSM) data to generate noisy pseudo ground truth of road network or building footprint and utilize it in the training process.

Our contributions are three-fold. First, we develop a novel method to generate pseudo road network and building footprint ground truth masks from OSM data without human annotation labor. Second, we propose a two-stage transfer learning to utilize the pseudo ground truth masks in the first stage, and fine-tune the model on high quality annotation data in the second stage. In addition, we also propose some techniques and architectural modifications, \ie, PP-LinkNet, to improve the overall performance of the binary semantic segmentation system to extract road networks and building footprints from satellite images. Finally, we achieve promising improvements on DeepGlobe road network dataset\cite{Demir2018}, SpaceNet road and building footprint dataset \cite{VanEtten2018}.

\begin{figure}[t!]
	\begin{center}
		\includegraphics[width=\linewidth,keepaspectratio]{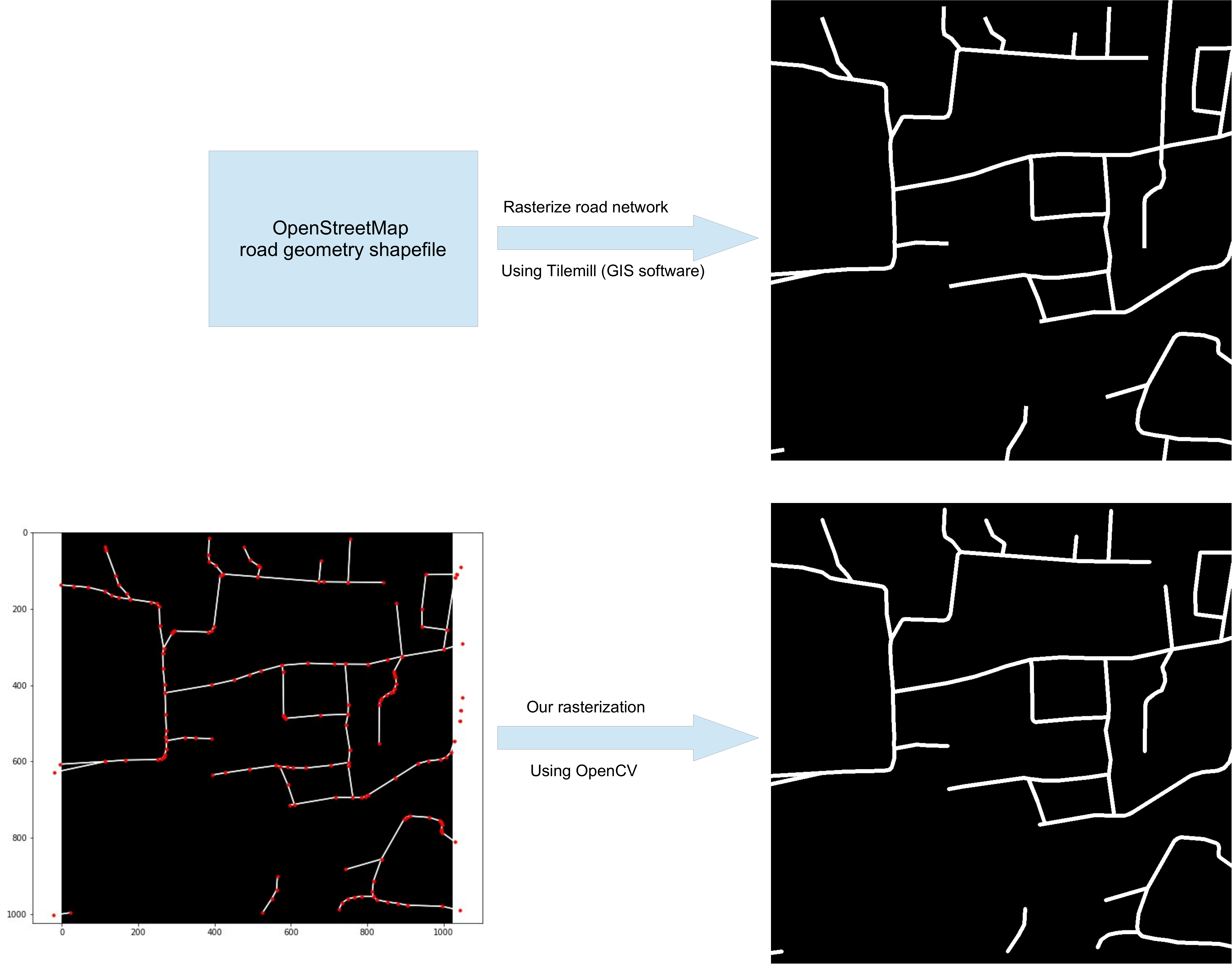}
	\end{center}
	\caption{Illustration of the rasterization process to convert OpenStreetMap data into pseudo ground truth mask image. Top row: rasterization using TileMill software. Bottom row: our rasterization using OpenCV from the corresponding road network graph. Best viewed in color.}
	\label{fig: rasterization process}
\end{figure}

\section{Related Work}
\textbf{Datasets for road network.} Two large datasets for road network extraction from satellite imagery (\ie, SpaceNet \cite{VanEtten2018} and DeepGlobe \cite{Demir2018}). They differ mainly in the format of annotation data -- while SpaceNet dataset provides centerlines for roads, DeepGlobe provides pixel labeling of roads. Several published works use proprietary datasets that are not available in the public domain for bench marking, e.g., \cite{MattyusRoadMaps,Mattyus_2017_ICCV,WangShenlong2017,Bastani2018}.

\textbf{Semantic segmentation models for road network extraction.} 
Inspired by successes of deep learning in computer vision \cite{He_CVPR2016,Ren2015,He2017,YeLuo_ICCV_2015,AnTran_ICCVW_2017}, there have been few attempts in addressing the problem of road extraction from satellite images using deep learning based semantic segmentation models. D-LinkNet \cite{Zhou2018}, the winner model of DeepGlobe road extraction challenge, utilizes LinkNet\cite{Chaurasia2018}, UNet\cite{Ronneberger2015}, ResNet \cite{He_CVPR2016} and dilated convolution\cite{Chen2016a} components model and  improvises using a post-processing step to refine the segmentation output to a road network graph. DeepRoadMapper \cite{Mattyus_2017_ICCV} developed a model based on ResNet \cite{He_CVPR2016} and FCN \cite{long_shelhamer_fcn}, and a technique to infer missing connections from segmentation results. On the other hand, RoadTracer \cite{Bastani2018} uses convolutional neural network (CNN) to predict the walking angle of the sliding window on the input image to decide on the input image for the next iteration.

Our work is similar to \cite{Audebert_2017_CVPR_Workshops}, where the authors use OSM data along with satellite imagery to predict semantic map features (\eg, surfaces, buildings, vegetables, trees, cars, etc). On the contrary, our work utilizes OSM as pseudo ground truth data only in the first stage of our two-stage transfer learning.

\section{Generating noisy pseudo ground truth}
\label{sect: pseudo ground truth}
In this section, we describe our approach to collect a large-scale satellite image dataset along with its pseudo ground truth to train road network and building footprint segmentation model in the first stage. 

\subsection{Rasterization of OSM data into pseudo (noisy) ground truth image}
In order to avoid laborious annotation work and to create a scalable approach towards creating training data we relied on OpenStreetMap (OSM)\cite{Hotosm}, a large community driven map. OSM data contains many useful metadata information such as type of roads, number of lanes, width, surface, bridge, tunnel, etc. Our source of satellite imagery is from different sources of remotely sensed data. The procedure to extract ground truth from OSM data is as follows: i) collect  satellite images for the area of interest, i) render the road-network from OSM using TileMill software. This rendering process is called the rasterization process in the geographical information system (GIS) community.  There are few GIS softwares that can do rasterization such as TileMill, QGIS, ArcGIS, \etc. Furthermore, we also develop our rasterization process using OpenCV. Specifically, our process is to draw lines with certain widths corresponding to locations of roads in the image. Figure~\ref{fig: rasterization process} shows that our rasterization process would produce similar pseudo ground truth image with the result from TileMill software. We use our rasterization process using OpenCV through all our experiments. Our rasterization process is based on the transformation matrix $\mathcal{T}$ available in tif-format satellite image, as in equation~\ref{equ:transformation matrix}),

\begin{equation}
(lon, lat) = \mathcal{T} * (r+0.5, c+0.5, 1),
\label{equ:transformation matrix}
\end{equation}
where $(lon, lat)$ is the longitude, latitude according to the image coordinate $(c, r)$. From equation~\ref{equ:transformation matrix}, given the longitude, latitude of a node in OSM road networks, we can infer its image coordinates from the inverse matrix of the transformation matrix $ \mathcal{T}$.

There are three main advantages of our approach of using noisy pseudo ground truth images as part of training data. First, annotating pixel-to-pixel for large image size of satellite images (\eg, $1024\times1024$) is an exhaustive task. Generating ground truth through rasterizing OSM data is efficient so that we can scale the number of training images up easily. Second, the pseudo ground truth of satellite imagery is generally available for most locations on earth. We only need to know the geographical address of the locations to rasterize the pseudo ground truth masks. Finally, the accuracy of OSM ground truth is enough to train a moderate model in the first phase. Hence, we devised a second phase in order to fine-tune our model with high quality annotated ground truth.

\begin{figure}[t]
	\begin{center}
		\includegraphics[width=\linewidth,keepaspectratio]{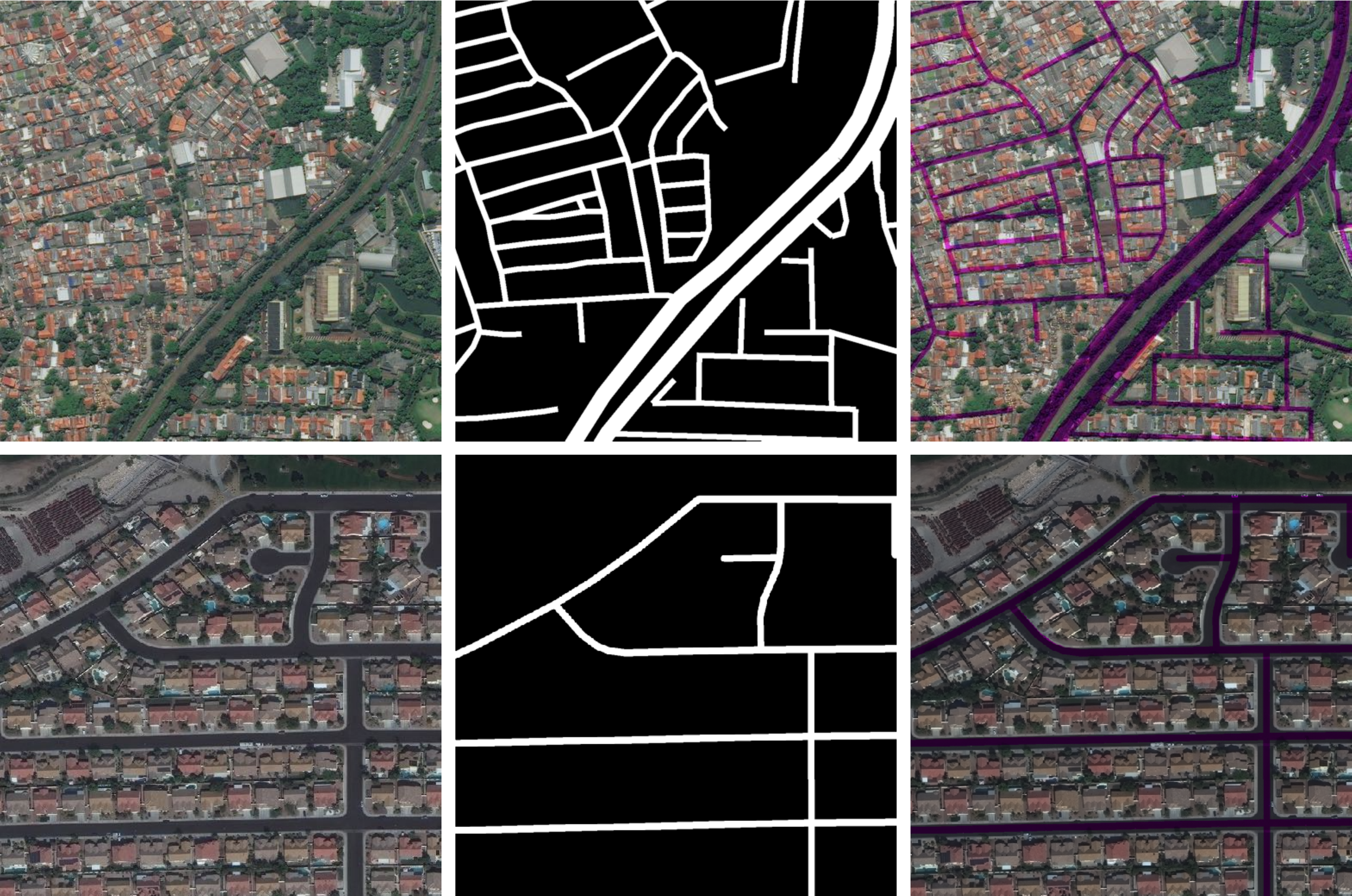}
	\end{center}
	\caption{Columns represent examples of inputs: satellite imagery, pseudo ground truth mask, and overlay of ground truth mask on input image. First row: a satellite image of \textit{Jakarta}; second row: a input image of \textit{Las Vegas} city in SpaceNet dataset.}
	\label{fig: rasterization output samples}
\end{figure}

Figure~\ref{fig: rasterization output samples} shows an example of a pair of input and its rasterized pseudo ground truth mask in Las Vegas city from SpaceNet dataset. Although the structure of road network in the example is correct with the satellite image, its pixel-to-pixel matching is limited. Sometimes, the pseudo ground truth is smaller or bigger than the actual road, hence, it is called pseudo ground truth.

\subsection{Image collection}
We build our training data in the first stage by combining data from SpaceNet \cite{VanEtten2018}, DeepGlobe \cite{Demir2018}, and our internal data. Both SpaceNet and DeepGlobe datasets have high resolution satellite images (\ie, 0.3-0.5m). We remove some imperfect images of SpaceNet \cite{VanEtten2018}, which has zero signal due to sensor's errors. The DeepGlobe \cite{Demir2018} images are sampled from DigitalGlobe Vivid imagery which contains cloud-free, high resolution satellite images. Furthermore, we also collect our own data from additional sources such as internal data, EarthExplorer \cite{USGS}, \textit{etc}. Currently, our dataset has two resolutions $1024\times1024$ and $1300\times1300$. We decide to build the dataset with high resolution satellite images because we consider the narrowness, connectivity, complexity, and long span of roads. Eventually, we utilize 48091 images and their corresponding noisy pseudo ground truth masks to train the model of road network segmentation in the first phase.

\section{Method}
We formulate our task of extracting roads from satellite images as a binary semantic image segmentation problem, where for each input satellite image, we predict if a pixel belongs to class 1 (Road) or class 0 (non-road). Distinct from datasets for object detection such as PASCAL VOC 2012 \cite{MVD2017} and MS-COCO \cite{Lin2014}, satellite images need to be processed at a high resolution (\eg, $1024\times1024$). This compels the need to design our network with efficient memory optimisations.

\subsection{Network Architecture Design}
\begin{figure}[t]
	\begin{center}
		\includegraphics[width=\linewidth,keepaspectratio]{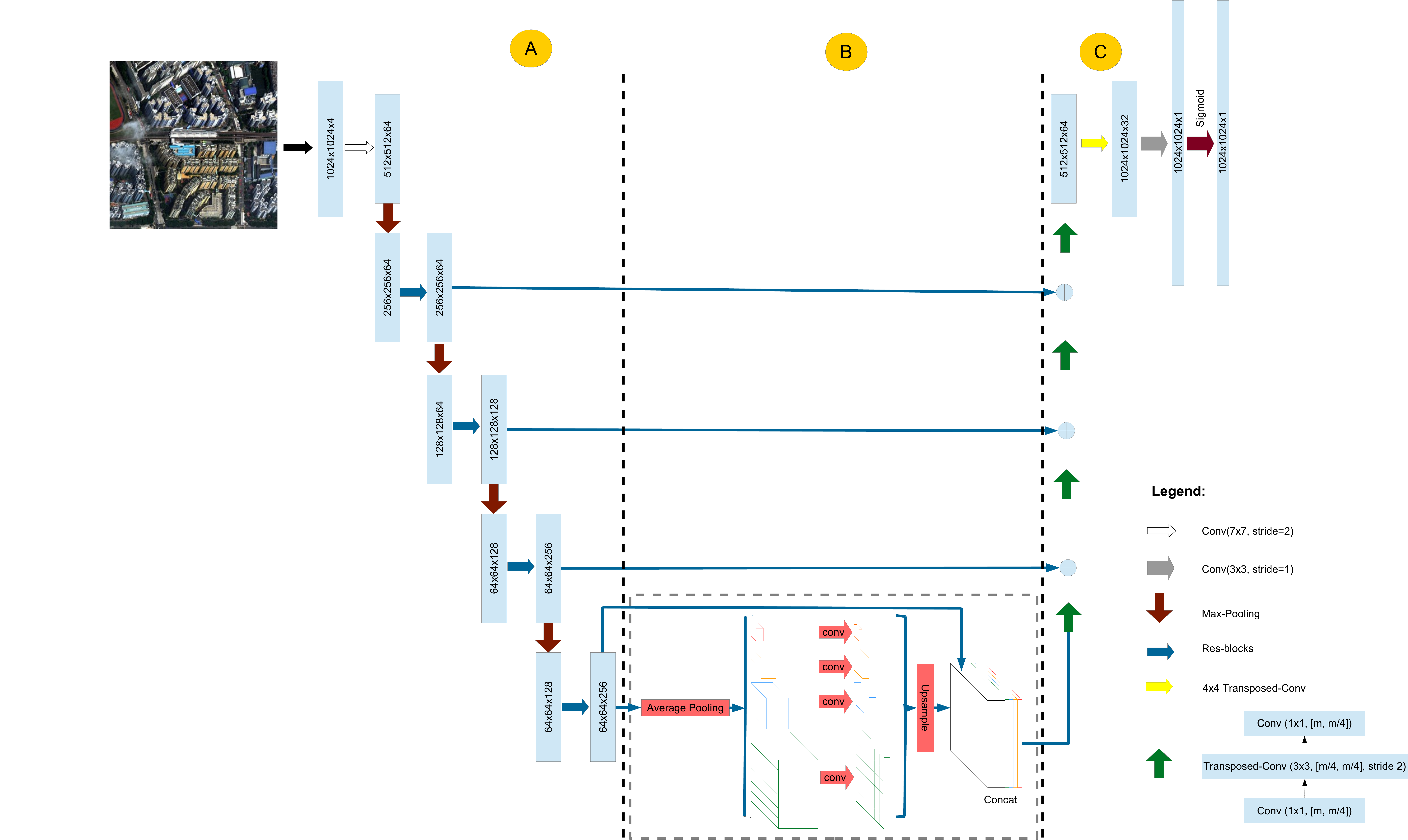}
	\end{center}
	\caption{Overview of our proposed network architecture PP-LinkNet. It is an improvement of D-LinkNet\cite{Zhou2018} with the replacement of enlarged receptive field module with pyramid pooling module in PSPNet \cite{Zhao2017}. Each blue block is a multi-dimensional feature which is an output from a block of convolutional layers (except the last layer is a sigmoid layer). Part A: encoder; part B: pyramid pooling module, part C: decoder. Best viewed in color.}
	\label{fig: network architecture}
\end{figure}

Figure~\ref{fig: network architecture} shows an overview of our proposed network architecture Pyramid Pooling-LinkNet (PP-LinkNet). Our network design is inspired by the encoder-decoder architecture such as Unet \cite{Ronneberger2015}. We received input images at resolutions of $1024\times1024$, hence, we emphasize on efficient network structure. We choose LinkNet \cite{Chaurasia2018} as our base network, similar to D-LinkNet \cite{Zhou2018}. Our encoder is ResNet34 \cite{He_CVPR2016} pre-trained on ImageNet dataset. RestNet34 is originally trained to classify images of mid-resolution $256\times256$, whereas, our task is to segment roads and building footprints from higher resolution imagery $1024\times1024$. Therefore, the network needs to adapt encoder layers to new domain of input. Our decoder includes bottleneck blocks, and layers of decoder up-sample feature size to have symmetric sizes with encoder's layers. The bottleneck block of a decoder consists of a transposed convolutional layer between two convolutional layers with $(1\times1)$-kernels (see Figure~\ref{fig: network architecture}). Our decoder is initialized with random parameters. In the middle of our network is global contextual prior, pyramid pooling (PP) module, which has been empirically proven to be effective in PSPNet \cite{Zhao2017}. Pyramid pooling module has no parameters, and proven to have better performance than the context module in D-LinkNet in our experiments.

\subsection{Improvements of semantic segmentation system for road network extraction}
\textbf{Two-stage transfer learning.} In the \textbf{first phase}, we train our PP-LinkNet on noisy pseudo ground truth generated from OSM (Sect.~\ref{sect: pseudo ground truth}) to let the network learn basic visual features and knowledge of the  new domain (\ie, aerial or satellite imagery).
For the \textbf{second phase}, we propose a simple transfer learning procedure to fine-tune the trained model one more time with high-quality well-annotated data.

\textbf{Class imbalance.} It is obvious that the number of non-road pixels is much bigger than the number of road pixels. This extreme class imbalance causes our network to pay more attention to background pixels to learn weights of semantic segmentation systems. Focal loss has been proposed in \cite{Lin2017} to tackle class imbalanced foreground and background sampled bounding boxes in object detection pipelines. Focal loss for each pixel $FL(p_{i,t})$ and for an image $FL$ can be expressed as
\begin{equation}
\begin{aligned}
FL(p_{i,t}) = -\alpha_{t}(1-p_{i,t})^{\gamma}\log(p_{i,t}) \\ 
FL = \sum_{i}^{N} FL(p_{i,t}),
\end{aligned}
\label{equ:focal loss}
\end{equation}
where $p_{i,t}$ is the model's estimated probability for the class with label $y=t$. Specifically, we have two class, non-road and road, $\{0, 1\}$ and $p\in[0, 1]$ is the estimated probability for the class with label $y=1$, then
\begin{equation}
p_t = \begin{cases} p, & \mbox{if y=1}  \\ 1-p, & \mbox{otherwise.} \end{cases}
\label{equ:pt definition}
\end{equation}
From different experiments, we found that $-\alpha_{t}, \gamma=0.5$ is the best parameters for the focal loss function $FL(p_{i,t})$.

Proposed in \cite{Milletari2016}, Dice coefficient loss, denoted as $DL$, can be written as
\begin{equation}
DL = 1 - \dfrac{2\sum_{i}^{N}p_i g_i + \epsilon}{\sum_{i}^{N}p_{i}^2 + \sum_{i}^{N}g_{i}^2 + \epsilon},
\label{equ:dice loss}
\end{equation}
where $N$ is the number of pixels in image, $p_i \in [0,1]$ is the predicted probability of a pixel, and $g_i \in \{0,1\}$ is the binary ground truth of the pixel. The $\epsilon$ term is a small constant to ensure the loss function stability by preventing numerical issues of dividing by 0. The Dice loss originated from Sørensen–Dice coefficient \cite{Dice_loss_wiki}, a statistic used for comparing similarity between two samples. Optimally, when we minimize $DL$ loss, an optimizer would increase $p_i$ for foreground pixels (\ie, $g_i=1$) and decrease $p_i$ for background pixels (\ie, $g_i=1$). Dice loss does not suffer from class imbalance because contributions of individual pixels only influence numerator or denominator term, not the whole loss function.

\textbf{Poly learning rate policy.} Similar to \cite{Chen2016a,Ramanathan2016}, we also explore a ``poly" learning rate policy (\ie, the learning rate is equal to the base learning rate multiplied by $(1 - \dfrac{iter}{max\_iter})^{power}$) with $power=0.9$ for training DLinkNet.

\textbf{Context module.} One improvement of PP-LinkNet compared to D-LinkNet is the pyramid pooling context module. We integrate the pyramid pooling module \cite{Zhao2017} into LinkNet \cite{Chaurasia2018} between encoder and decoder part. Through our empirical experiments, we found that the pyramid pooling module with no parameters is effective in improving performance of our model, hence, we named it PP-LinkNet (\ie, PyramidPooling-LinkNet).

\begin{figure}[t]
	\centering
	\begin{subfigure}[t]{0.5\linewidth}
		\centering
		\includegraphics[width=\linewidth,keepaspectratio]{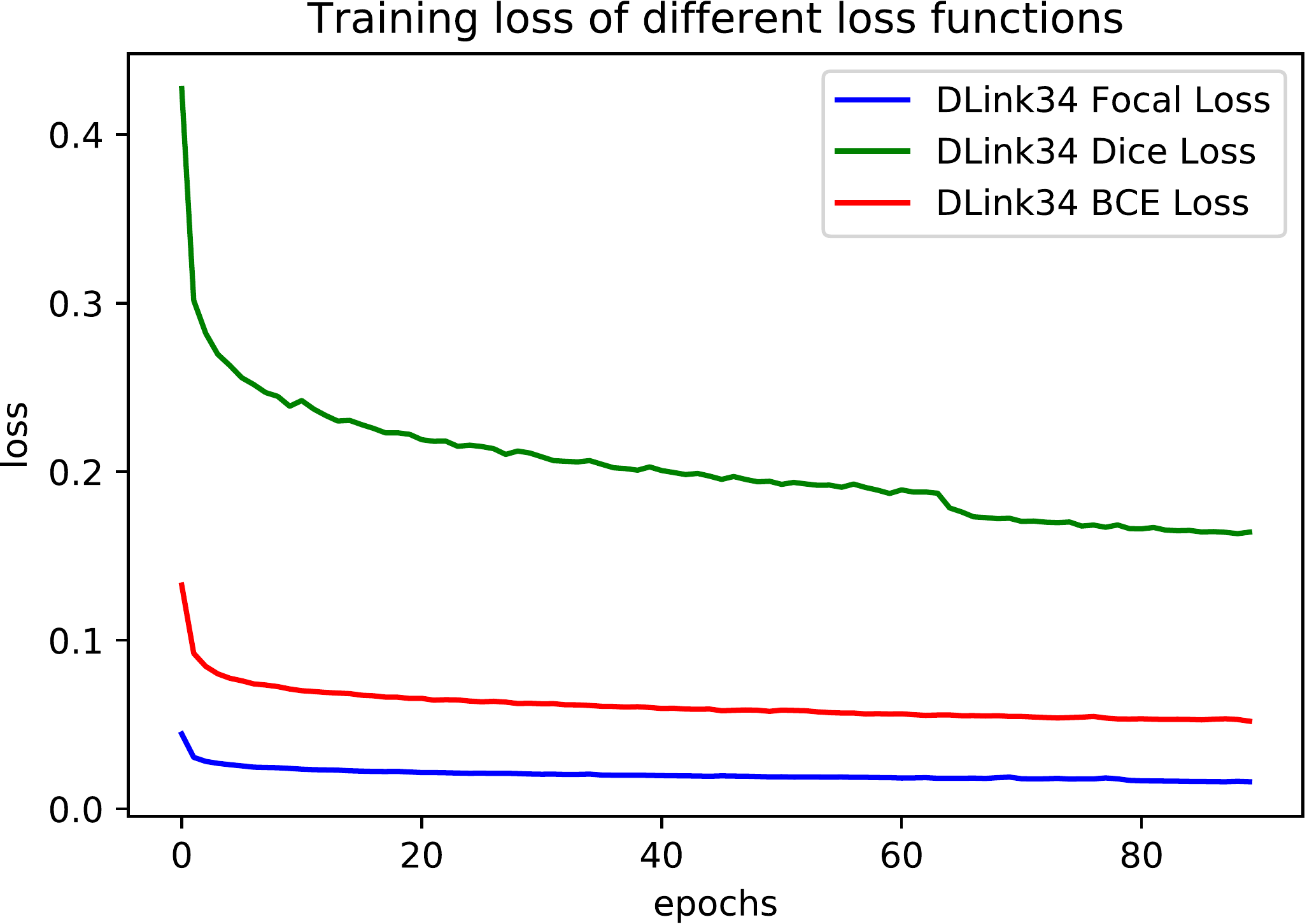}
		\caption{Three loss functions}
		\label{fig: three loss functions}
	\end{subfigure}%
	~
	\begin{subfigure}[t]{0.5\linewidth}
		\centering
		\includegraphics[width=\linewidth,keepaspectratio]{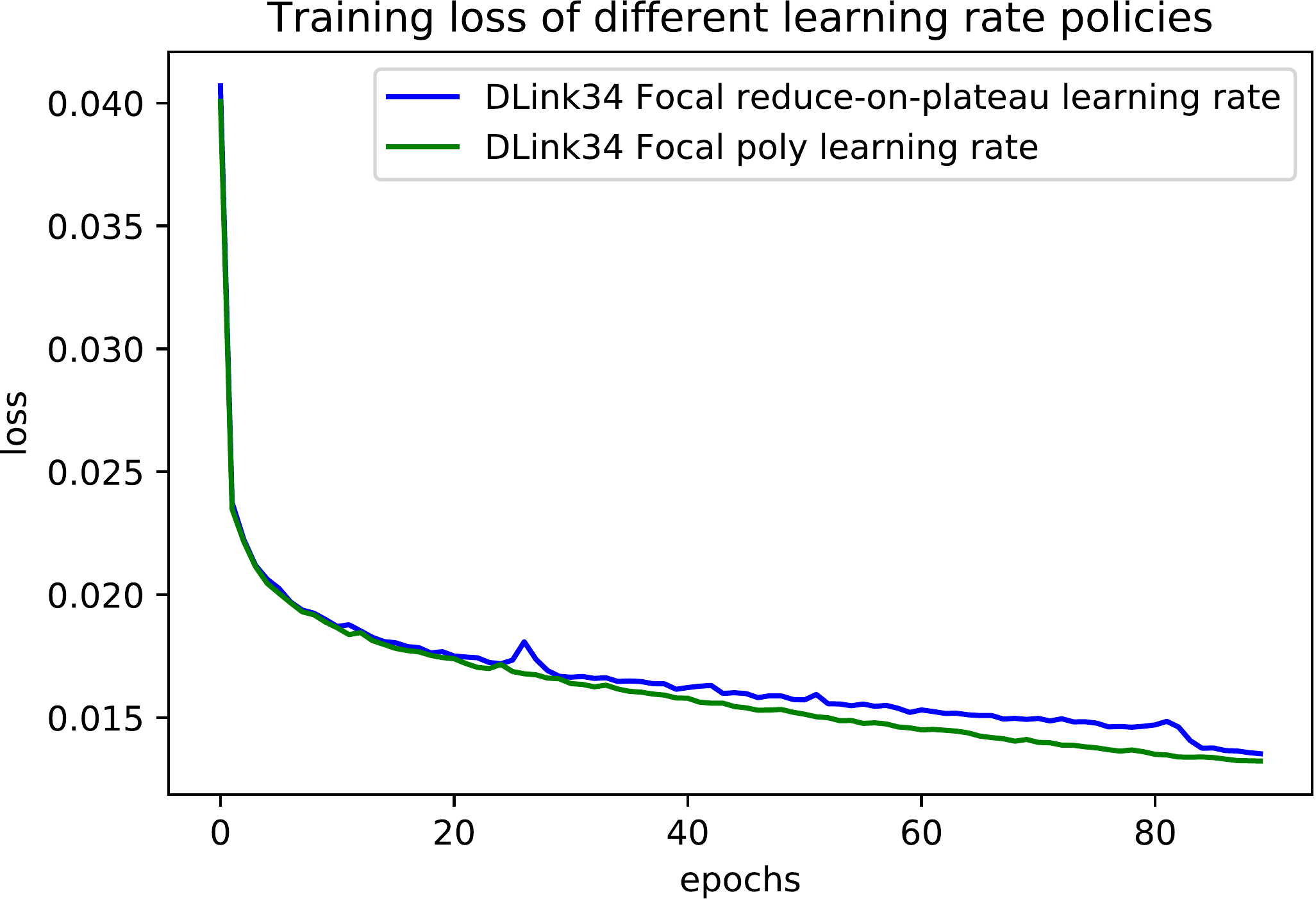}
		\caption{Effects of poly learning rate}
		\label{fig: poly lr}
	\end{subfigure}
	
	\caption{$\left( a \right)$. Training loss of three different loss functions: focal loss, dice loss, binary cross entropy loss. $\left( b \right)$ Comparison of training losses between reduce-on-plateau and poly learning rate policy. The experiments are done with the same settings: data, batch size, learning rate. The loss values are averaged per epoch.}
	\label{fig: three loss functions and poly learning rate}
\end{figure}

\section{Experiments}
\subsection{Dataset}
\textbf{DeepGlobe road extraction \cite{Demir2018}.} Road extraction task in DeepGlobe challenge has 6226 training, 1243 validation, and 1101 testing images, with image resolution of $1024\times1024$. The ground resolution is 50 cm/pixel. However, we don't have annotations for validation and testing images. We follow \cite{Singh2018} to split the training image set into 4696 images for training and 1530 for validation.

\textbf{SpaceNet challenge 2 building footprints \cite{Demir2018}.} The dataset includes 24,586 images of size $650\times650$ containing 302,701 building footprints across Las Vegas, Paris, Shanghai, Khartoum. The ground resolution is 30 cm/pixel. We split the training set of dataset into 80\%/20\% for training/test, equivalently to, 8472 images for training and 2121 for testing. We conduct experiments on only 3 RGB channels of multi-spectral images, and on full resolution $650\times650$ of input images. More information about F1 evaluation metric of the dataset can be found in DeepGlobe building detection task \cite{Demir2018}.

\textbf{SpaceNet challenge 3 road extraction \cite{Demir2018}.} The dataset includes 2780 images, and, following \cite{Singh2018}, we split the dataset into 2213 images for training and 567 images for testing. The ground truth is represented by line strings, indicating center-lines of the roads. The image has a resolution of $1300\times1300$, and its ground resolution is 30cm/pixel. Unlike \cite{Batra_2019_CVPR}, we train our model on larger image crop $1024\times1024$, not $512\times512$. As human errors, we always have some mistakes in annotations of three datasets, but we leave it for future analyses.

\subsection{Goal and Metrics}
\textbf{Pixel based metric.} The semantic segmentation algorithm is expected to predict a mask for a test image (\ie, road and non-road labels). Similar to \cite{Demir2018}, we deploy the standard Jaccard Index (\ie, the PASCAL VOC intersection-over-union IoU metric \cite{Everingham10}), defined as Equ.~\ref{equ:IoU formula},
\begin{equation}
IoU_i = \dfrac{TP_i}{TP_i + FP_i + FN_i},
\label{equ:IoU formula}
\end{equation}
where $TP_i$ is the number of pixels that are correctly predicted as road pixel, $FP_i$ is the number of pixels that are wrongly predicted as road pixel, and $FN_i$ is the number of pixels that are wrongly predicted as non-road pixel for image $i$. The final evaluation score is computed as the mean IoU of all $n$ images (\ie, Equ.~\ref{equ:mIoU formula}),
\begin{equation}
mIoU = \dfrac{1}{n}\sum_{i=1}^{n}IoU_i.
\label{equ:mIoU formula}
\end{equation}

\textbf{Graph based metric.} To measure the difference between estimated road network and ground truth graph, we use Average Path Length Similarity (APLS) in \cite{VanEtten2018}, as evaluation metric. The APLS metric sums up the differences in optimal path lengths between nodes in the estimated graph $G'$ and the ground truth graph $G$. The ALPS metric scales from 0 (poor) to 1 (perfect), 
\begin{equation}
C = 1 - \frac{1}{N} \sum \min \left\lbrace 1, \frac{|L(a, b) - L(a', b')|}{L(a, b)} \right\rbrace
\label{equ: APLS metric}
\end{equation}
where $N$, number of unique paths, while, $L(a, b)$, length of path $(a, b)$. The node $a'$ denotes the projected node (\ie, into the ground truth graph) in the proposal graph closest to the node $a$ of the ground truth graph. We also have a similar metric from the ground truth graph $G$ into the proposal graph $G'$.  If path $(a', b')$ does not exist, then the maximum penalty would be used, 1. Hence, a missing or redundant segment would also be penalized.

\begin{figure}[t]
	\centering
	\begin{subfigure}[t]{0.5\linewidth}
		\centering
		\includegraphics[width=\linewidth,keepaspectratio]{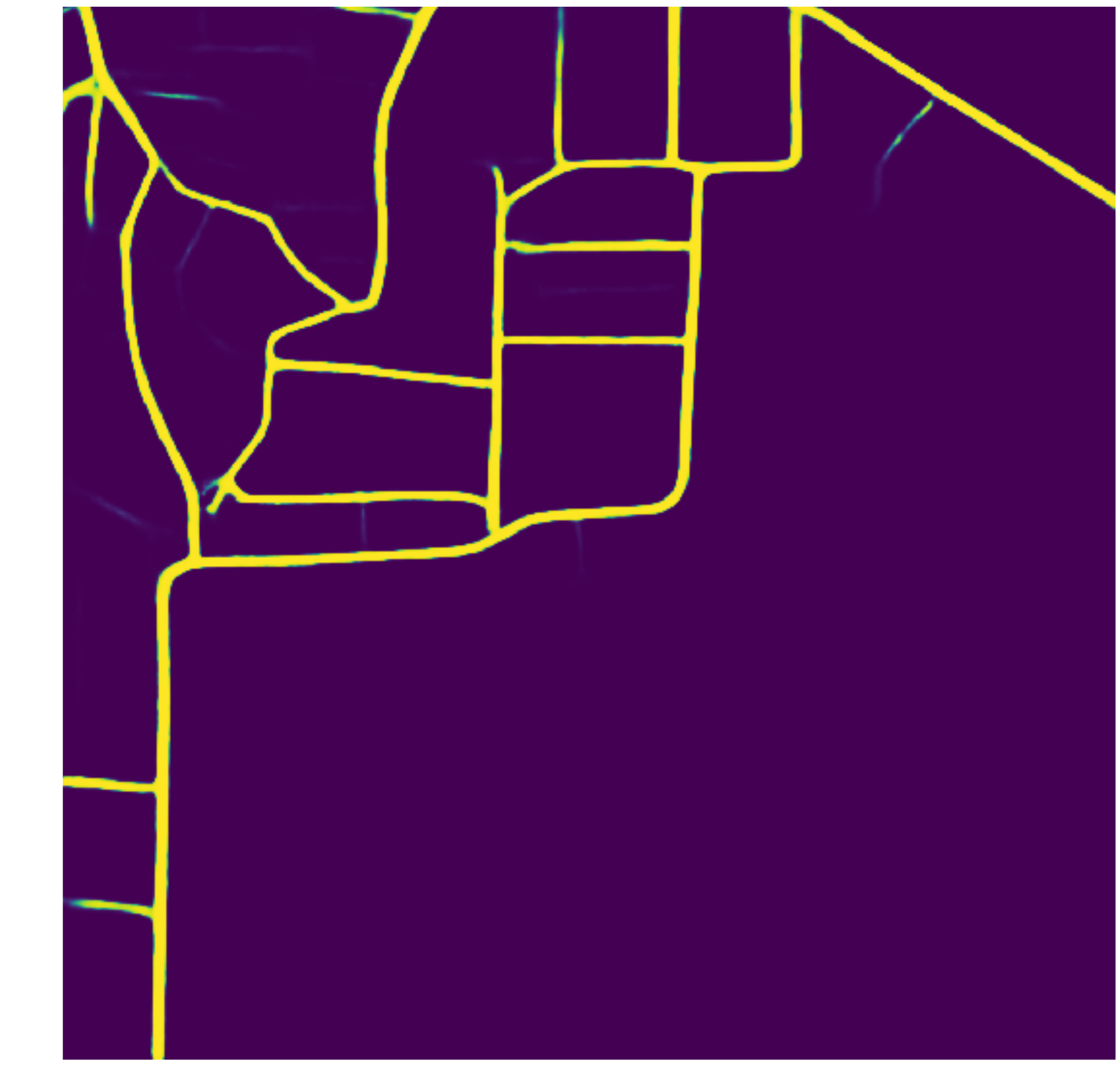}
		\caption{BCE+Dice loss confidence map}
	\end{subfigure}%
	~
	\begin{subfigure}[t]{0.5\linewidth}
		\centering
		\includegraphics[width=\linewidth,keepaspectratio]{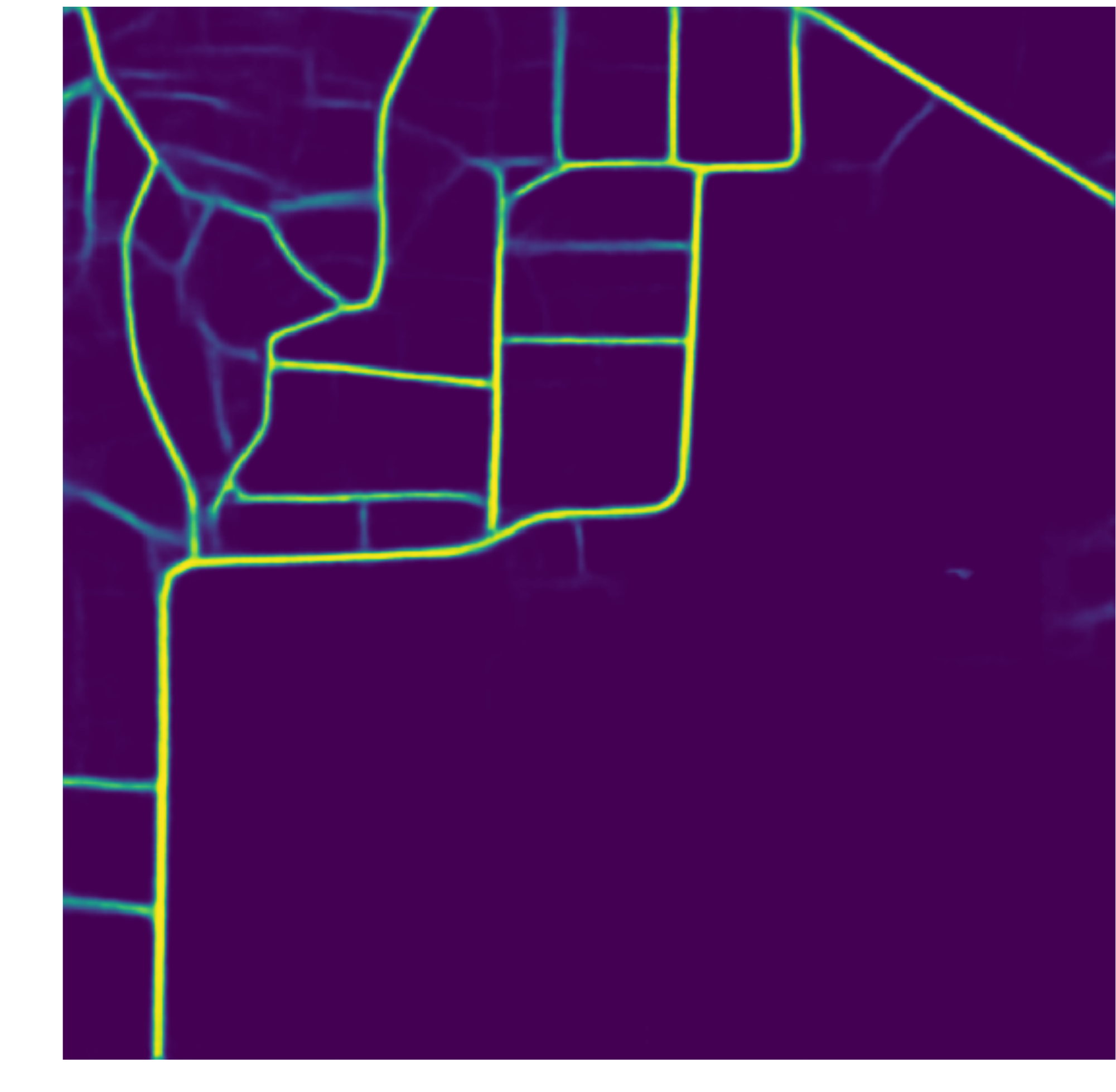}
		\caption{Focal loss confidence map}
	\end{subfigure}
	\caption{Illustration of confidence map of the BCE+dice loss model and focal loss model on an example image. Warm color indicates high confidence value. Best viewed in color.}
	\label{fig:confidence map of bce_dice and focal loss example}
\end{figure}

\subsection{Implementation Details}
\textbf{Baseline models.}  We compare our method with recent state-of-the-art methods that are all based on deep learning and leverage pre-trained models on large-scale image classification dataset such as ImageNet \cite{imagenet_cvpr09}. There are several algorithms proposed for semantic segmentation such as FCN \cite{long_shelhamer_fcn}, U-Net \cite{Demir2018}, DeepLab \cite{Chen2016a}, SegNet \cite{Badrinarayanan2015}, PSPNet \cite{Zhao2017}, LinkNet \cite{Chaurasia2018}, ENet \cite{Paszke2016}, \textit{etc}. However, there are few works adapted for road network extraction from satellite images. In this work, we present our baseline results based on three networks TernausNetV2 \cite{Iglovikov2018}, PSPNet \cite{Zhao2017}, and D-LinkNet \cite{Zhou2018}.

TernausNetV2 \cite{Iglovikov2018} and D-LinkNet \cite{Zhou2018} are the winner models for building detection task, and road extraction task respectively in the DeepGlobe challenge 2018 \cite{Demir2018}. Both tasks share similar characteristics which are dense prediction of per-pixel labels (\ie, building polygons and road surface). An improved version of PSPNet \cite{Zhao2017} is the winning submission of the 2017 Large Scale Scene Understanding (LSUN) challenge on scene semantic segmentation.

\textbf{Training procedures.\label{sect: training}} There are interesting evidences that a bigger crop size of input image would boost performance of semantic segmentation problem \cite{Bulo2017,Wang2018}. Our PP-LinkNet is designed for dealing with high resolution satellite images of $1024\times1024$ pixels. In order to have fair comparisons, we also train three baseline algorithms with the same resolution of input images. For the three baseline algorithms, we follow training as detailed by the authors. For example, for D-LinkNet \cite{Zhou2018}, we use Adam optimizer with learning rate $2e^{-4}$, and reduced by 5 for 3 times when training loss decreases slowly. We train our PP-LinkNet in 90 epochs with learning rate $2e^{-4}$ and batch size of 9 in our experiments. TernausNetV2 consumes large GPU memory for a big crop size of $1024\times1024$, hence, we only train TernausNetV2 networks with half-precision floating-point numbers 16-bits \footnote{\url{https://github.com/NVIDIA/apex}} \cite{Micikevicius2017} (\ie, fp16) and a batch size of 2.

\subsection{Results}
\textbf{Ablation study.} Table~\ref{tab: ablation results} shows the ablation study for our PP-LinkNet designs and the two-stage transfer learning. As can be observed, both DLinkNet34 and PSPNet outperforms TernausNetV2 fp16. It might be due to the fact that the mixed precision training is not yet standardized. DLinkNet34 has better performance than PSPNet (67.42\% vs. 63.38\% mIOU), although DLinkNet34 and PSPNet have the same ResNet34 backbone network.

\begin{table}[t]
	\begin{center}
		\begin{tabular}{|l|c|c|}
			\hline
			Improvement & mean IoU & APLS score \\
			 &  (\%) &  (\%) \\
			\hline\hline
			TernausNetV2 fp16 \cite{Iglovikov2018} 		& 47.16  & N/A \\
			PSPNet \cite{Zhao2017} 								& 63.38  & N/A \\
			DLinkNet34 \cite{Zhou2018} 						 & 67.42  & 71.85 \\
			\hline
			DLinkNet34 + Dice loss		& 66.31 & N/A \\
			DLinkNet34 + bce loss		& 67.21 & N/A \\
			DLinkNet34 + focal loss		& 67.82 & N/A \\
			\hline
			DLinkNet34 + focal + poly lr												 & 69.49 & 75.71 \\
			PP-LinkNet34 (Ours)	 	   					& 69.87 & 76.04 \\
			\hline
			PP-LinkNet34 first-stage (on OSM data)									  & 68.12 & 73.77 \\
			PP-LinkNet34 two-stage (Ours)											& \textbf{70.53} & \textbf{77.11} \\
			
			\hline
		\end{tabular}
	\end{center}
	\caption{Ablation study of different design choices for our PP-LinkNet architecture with two-stage training on DeepGlobe road extraction dataset. First block: baseline results; second block: various modifications on D-LinkNet34; third block: evolution of PP-LinkNet; forth block: comparisons between two-stage vs. one-stage training.}
	\label{tab: ablation results}
\end{table}

Figure~\ref{fig: three loss functions} shows the training behavior of three loss functions. Interestingly, as can be seen, the focal loss obtains the smallest training loss compared to the BCE and Dice loss. The Dice loss is harder than the BCE and focal loss to be optimized in DLinkNet framework, but it also shows a monotonically decreasing loss values. Table~\ref{tab: ablation results} presents meanIOU and APLS scores on test set of Deepglobe road extraction dataset \cite{Demir2018}. It is consistent with the training loss behaviours that our result for the focal loss model (\ie, 67.82\%) is higher the BCE (\ie, 67.21\%), and Dice loss model (\ie, 66.31\%) respectively. Figure~\ref{fig:confidence map of bce_dice and focal loss example} illustrates the difference in confidence map of the BCE+Dice loss model compared to the focal loss model. With the same input image, the BCE+Dice model pays attention to main roads, while the focal loss model is able to put confidence on both main and small roads. This observation is also reflected in APLS scores with a big margin. Our final model with focal loss has improved on the original DLinkNet34\cite{Zhou2018} with BCE+Dice loss 5.26\% on APLS scores (\ie, 77.11\% vs. 71.85\%). We also argue that APLS score is more important meanIOU score, because it is inherent to connectivity of road network graphs. From following experiments, we use the focal loss for training PP-LinkNet because it is shown as a single effective loss function for the road extraction task.

Figure~\ref{fig: poly lr} shows the training behavior of the poly learning rate policy. The poly learning rate policy is more effective to optimize the deep semantic segmentation system than the reduce-on-plateau learning rate policy. As a result, we improve DLinkNet34 with poly learning rate 1.67\% meanIoU from $67.82\%$ to $69.49\%$. We will use the poly learning rate for the following experiments.

\begin{figure}[t]
	\begin{center}
		\includegraphics[width=\linewidth,keepaspectratio]{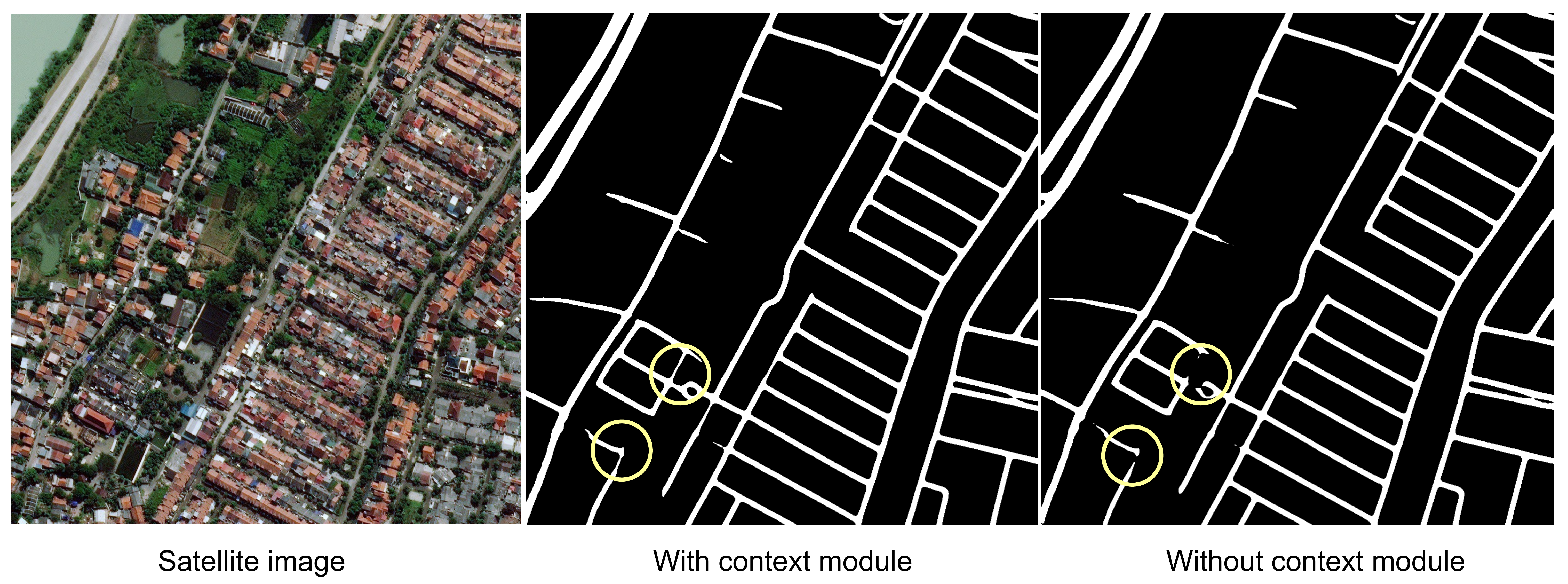}
	\end{center}
	\caption{The impact of the pyramid pooling context module in PP-LinkNet. Left: an input image; middle: the road segmentation mask with PP-Linknet; right: the segmentation mask without the context module in PP-LinkNet. Best viewed in color.}
	\label{fig: context module impacts}
\end{figure}

Additionally, Table~\ref{tab: ablation results} shows the result of PP-LinkNet34 model with pyramid pooling (PP) context module is better than D-LinkNet34 model with dilation context module (\ie, 69.87\% vs. 69.49\% in meanIoU, 76.04\% vs. 75.71\% in APLS). Compared to the original D-LinkNet34, we obtain a larger margin of performance (\ie, 69.87\% vs. 67.42\% in meanIoU, 76.04\% vs. 71.85\%) and it suggests that other proposed modifications also have compound effects with the pyramid pooling context module. In other experiments in SpaceNet road network and building footprint dataset (see Table~\ref{tab: road extraction results},~\ref{tab: SpaceNet building footprint extraction}), our PP context module also has consistently better performance than dilation context module. Figure~\ref{fig: context module impacts} shows that our context module can help PP-LinkNet to infer road segments that are difficult to predict when PP-LinkNet does not have the context module.

Finally, our two-stage training further boost performance of the network especially in terms of APLS scores. PP-LinkNet two-stage improves on the first-stage one 3.34\% in APLS scores (\ie, 77.11\% vs. 73.77\%), and also achieve the best results over all model variations.

\begin{table}[t]
	\begin{center}
		\resizebox{\linewidth}{!}{
		\begin{tabular}{|l|c|c|}
			\hline
			\textbf{Method} 									& \textbf{SpaceNet} 	&  \textbf{DeepGlobe}  \\
			& mIOU \quad APLS			& mIOU \quad APLS \\
			\hline\hline
			DeepRoadMapper (segmentation) \cite{Mattyus_2017_ICCV} & 59.99 \quad 54.25 & 62.58 \quad 65.56 \\
			DeepRoadMapper (full) \cite{Mattyus_2017_ICCV} 		 & N/A \quad 50.59 		  & N/A \quad 61.66 \\
			RoadCNN (segmentation) \cite{Bastani2018}   		   & 62.34 \quad 58.41     & 67.61 \quad 69.65 \\
			MatAN \cite{Mattyus_2018_CVPR}								& 52.86 \quad 46.44     & 46.88 \quad 47.15 \\
			Topology Loss (with BCE) \cite{Mosinska2017}		  & 56.29 \quad 49.00		& 64.95 \quad 56.91 \\
			Topology Loss (with SoftIoU) \cite{Mosinska2017}	& 57.69 \quad 51.99			& 64.94 \quad 65.96 \\
			LinkNet34 \cite{Chaurasia2018}			   						& 60.33 \quad 55.69		& 62.75 \quad 65.33 \\
			D-LinkNet34 \cite{Zhou2018}	(Ours reproduced)		& 55.50 \quad 64.74	& 67.42 \quad 71.86 \\
			Improved road connectivity \cite{Batra_2019_CVPR}   & 63.75 \quad 63.65 	& 67.21 \quad 73.12 \\
			Self-supervision Coach Mask	\cite{Singh2018} 			& \textbf{77.0}	\quad	N/A	 	  & \textbf{76.8} \quad N/A \\
			Sat2Graph \cite{He2020}													 & N/A \quad 64.43 & N/A \quad N/A \\
			\hline
			PP-LinkNet34 (Ours)								& 55.50 \quad 65.94 & 69.87 \quad 76.04  \\
			PP-LinkNet34 two-stage (Ours) 			 & 56.43 \quad \textbf{67.03} & 70.53 \quad \textbf{77.11} \\
			\hline
		\end{tabular}
	}
	\end{center}
	\caption{Our results of our PP-LinkNet models on the road network extraction task, and comparison of our technique with the state-of-the-art road network extraction techniques on SpaceNet and DeepGlobe road extraction dataset.}
	\label{tab: road extraction results}
\end{table}

\begin{table}[t]
	\begin{center}
		\begin{tabular}{|l|c|c|}
			\hline
			Method & meanIoU 		  &  F1 score \\
						 &  (\%) 				&  (\%) \\
			\hline\hline
			PP-LinkNet34 two-stage (Ours) 			 &  \textbf{78.19} & \textbf{68.92} \\
			PP-LinkNet34 first-stage (OSM data)	   & 44.00 & 25.33 \\
			PP-LinkNet34 (Ours)								& 76.62 & 67.04 \\
			\hline
			DLinkNet34 \cite{Zhou2018} (Ours reproduced)	&  76.21  &  66.79  \\
			\hline
		\end{tabular}
	\end{center}
	\caption{Results of our PP-LinkNet model on building footprint detection task. Top half: different variations of our PP-LinkNet; Bottom half: our reproduced results from other works.}
	\label{tab: SpaceNet building footprint extraction}
\end{table}
\footnotetext{The model has been tested on a subset of training set. Our train and test set is split from the original training set of the model.}

\textbf{Road network extraction.} Table~\ref{tab: road extraction results} shows our results on SpaceNet and DeepGlobe road extraction dataset. With our two-stage training, we are able to improve quantitative metrics over PP-LinkNet one-stage training over two datasets. We achieve larger gaps of improvements in APLS scores in both dataset. APLS score involves topological measures, hence, it means that our two-stage approach boosts the ability of PP-LinkNet to detect road connectivity.

Furthermore, Table~\ref{tab: road extraction results} also shows that our PP-LinkNet obtains the new state-of-the-art performance on both SpaceNet and DeepGlobe dataset. Notice that SpaceNet road dataset does not have pixel-level annotations, hence, its meanIOU evaluation based on the assumption that all roads have a fixed road width, which is often not true. Batra \etal \cite{Batra_2019_CVPR} proposed a method to learn road orientation at the same time with segmentation, and used it to refine connectivity of roads. Despite that, we outperform \cite{Batra_2019_CVPR} with regard to APLS scores in both dataset (\ie, 3.38\% in SpaceNet (67.03\% vs. 63.65\%), 3.99\% in DeepGlobe (77.11\% vs. 73.12\%)). We also outperform other popular techniques for road network extraction such as DeepRoadMapper\cite{Mattyus_2017_ICCV}, D-LinkNet\cite{Zhou2018}, RoadTracer\cite{Bastani2018} \etc. Furthermore, our result is also better than the recent Sat2Graph \cite{He2020} technique on SpaceNet dataset.

\begin{figure}[t]
	\centering
	\includegraphics[width=\linewidth]{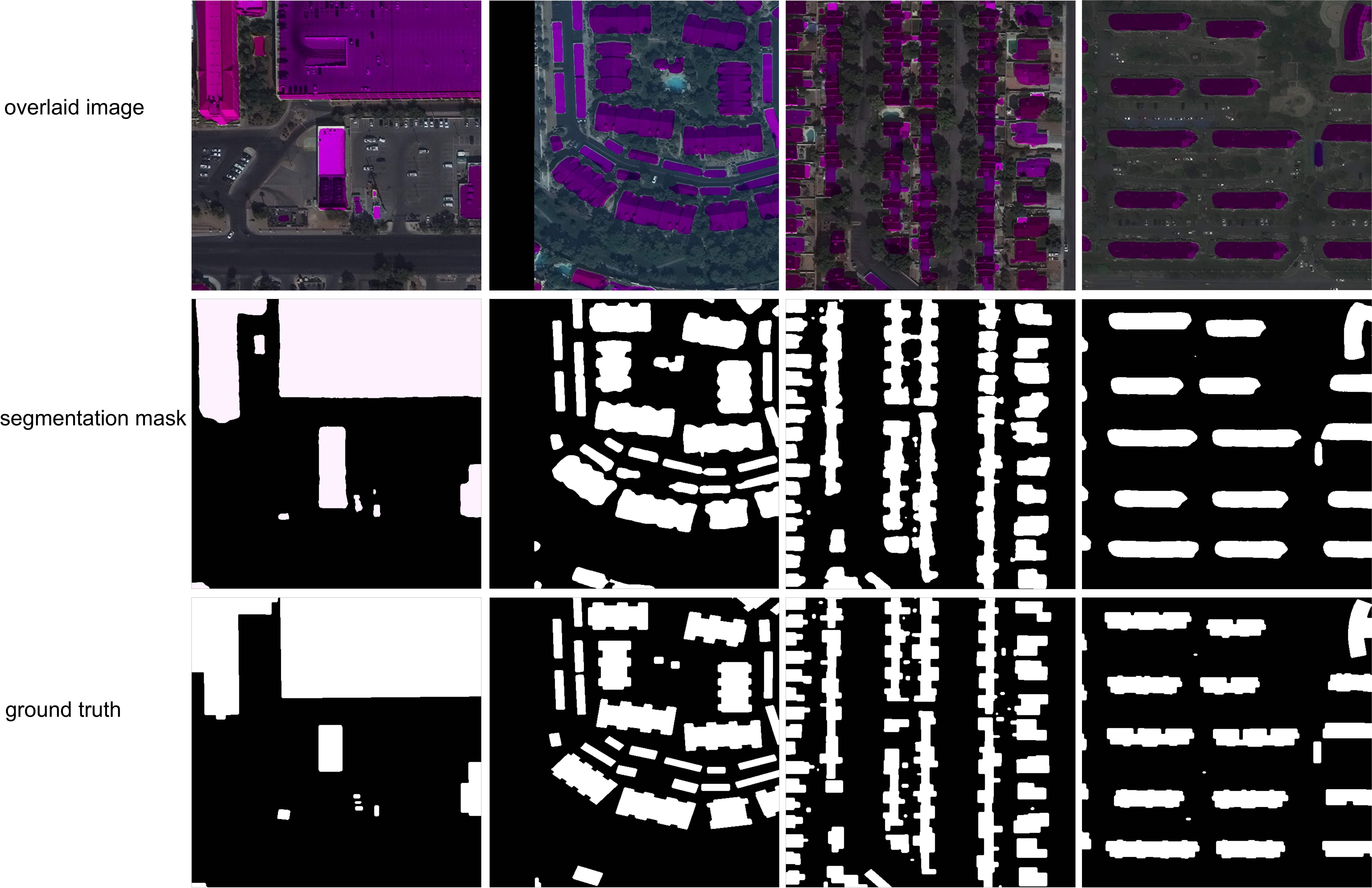}
	\caption{Examples of building footprint extraction from SpaceNet building dataset. We only use 3 channels R, G, B among multi-spectral imagery to train and test our model. From top to bottom:  overlaid image between segmentation mask and RGB input image, segmentation mask, and ground truth image.}
	\label{fig: examples of extracted building footprints}
\end{figure}

\textbf{Building footprint extraction.} Table~\ref{tab: SpaceNet building footprint extraction} shows our results on building footprint extraction tasks. As can be observed, our PP-LinkNet two-stage out-performs all other variations in both \textit{meanIOU} and \textit{F1 score} at IOU of 0.5. Similarly to the road extraction problem, our PP-LinkNet first-stage does not show good quantitative results because the rasterized OSM pseudo ground truth is not accurate pixel-to-pixel matching between mask and satellite imagery. Our PP-LinkNet two-stage also performs better than PP-LinkNet one-stage (\ie, 78.19\% vs. 76.62\% in meanIoU, 68.92\% \vs 67.04\% in F1 score). Our work is the first work that evaluates our models on SpaceNet building footprint dataset. Hence, it is hard to compare with other existing models \cite{Hamaguchi_2018_CVPR_Workshops,Golovanov2018,Iglovikov2018}. We have better performance than our reproduced DLinkNet34 \cite{Zhou2018}, and it indicates that pyramid pooling module captures global scene context better than DLinkNet34's context module. We don't have training script for TernausNetV2 \cite{Iglovikov2018} trained on both our train and test set, therefore, its released model's result would approximate the upper-bound performance on the dataset (\ie, 76.76\% in F1 score). Furthermore, our PP-LinkNet model has performance on par with the winner of SpaceNet building detection (\ie, F1=69\%), although we don't use multi-spectral channels and ensembles of multiple deep learning models. Figure~\ref{fig: examples of extracted building footprints} demonstrates some visualizations of our segmentation result mask. It shows that our model can pick out building footprint reasonably well (\ie, third column in Figure~\ref{fig: examples of extracted building footprints}).

\begin{figure}[t!]
	\centering
	\begin{subfigure}[t]{0.5\linewidth}
		\centering
		\includegraphics[width=\linewidth,keepaspectratio]{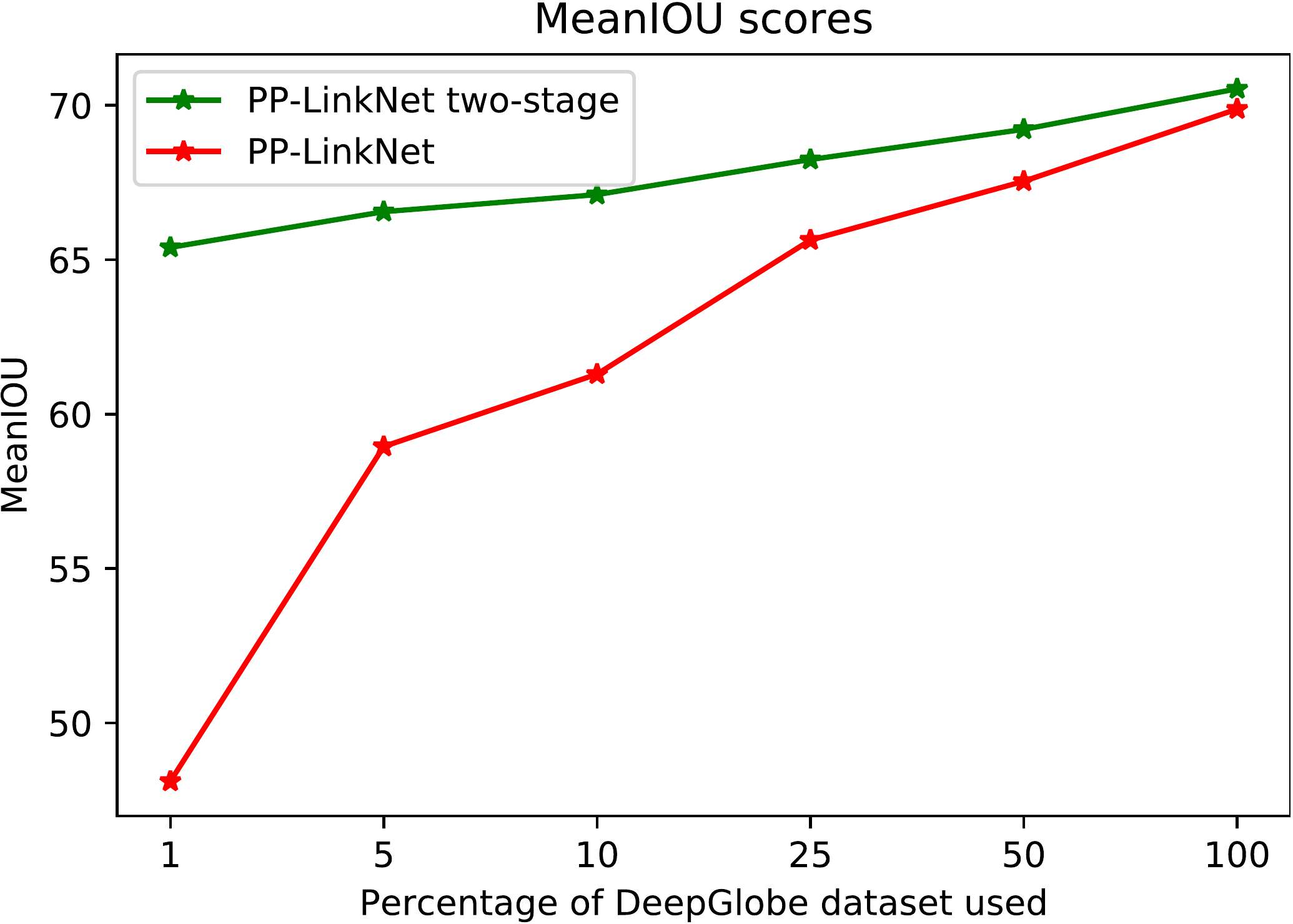}
		\caption{MeanIOU scores}
	\end{subfigure}%
	~
	\begin{subfigure}[t]{0.5\linewidth}
		\centering
		\includegraphics[width=\linewidth,keepaspectratio]{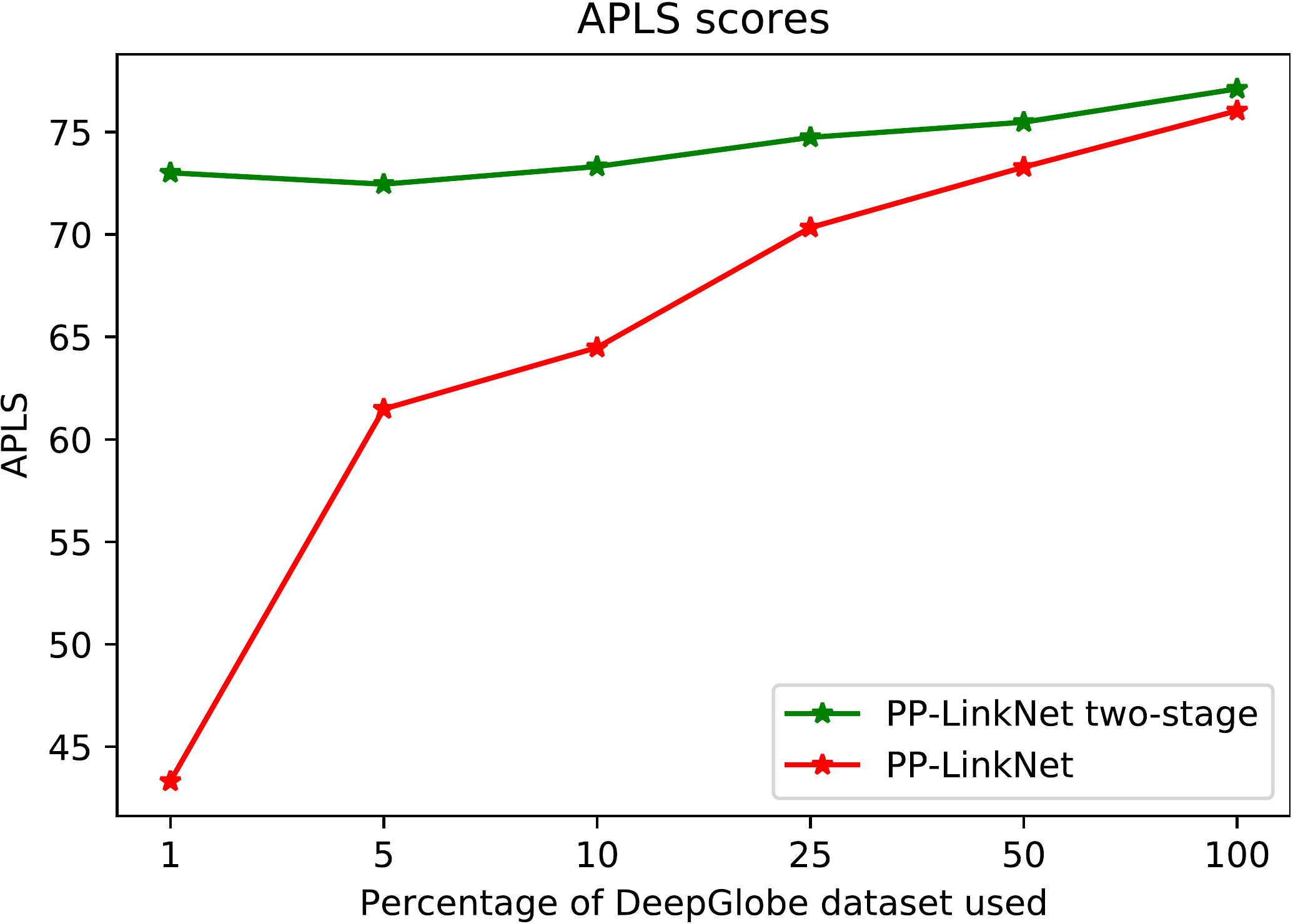}
		\caption{APLS scores}
	\end{subfigure}
	\caption{Performance of PP-LinkNet two-stage training compared with PP-LinkNet in term of meanIOU and APLS score with respect to the fraction of DeepGlobe road dataset used for training the model. Best viewed in color.}
	\label{fig: fine-tuning on small data}
\end{figure}

\textbf{Transfer learning on small data.} Figure~\ref{fig: fine-tuning on small data} demonstrates the effects of our two-stage transfer learning for semantic segmentation in the road extraction task. We vary the fraction of DeepGlobe road extraction dataset for training our PP-LinkNet model by selecting randomly 1\%(\ie, equivalently to 46 samples), 5\%, 10\%,  25\%, 50\% and 100\% subset of the dataset. Our PP-LinkNet two-stage performance is better than PP-LinkNet by a large margin in the extremely low number of training samples (\eg, 1\%, 5\%). The gap of performances is gradually reduced when we have more training samples. In this experiment, we show that our two-stage training procedure helps transfer learning to be more effective in learning visual representations from a small number of satellite images.

\section{Discussions}

Domain adaptation is an important procedure for adapting the models of supervised learning into domains different with the trained environment. Often, the new domain (\eg, aerial, medical imagery) has small data or small annotated data. Our two-stage transfer learning is beneficial in adapting supervised learning trained with big data into new domains with small data.

Our two-stage transfer learning is also closely related to training deep learning models in the revival era of deep learning \cite{Hinton:2006:FLA:1161603.1161605}. At that time, we pre-train each layer of Restricted Boltzmann Machine in unsupervised fashion, and then, we fine-tune the whole network for different tasks. In our work, with the prevalence of supervised learning techniques, computing resources, and data, we first pre-train the whole deep model on pseudo-groundtruth (noisy) data, then fine-tune it on high-quality annotated data.

\section{Conclusion}
This paper introduces a new procedure to generate and utilize per-pixel pseudo ground truth labels for extracting road network and building footprint in satellite images. The procedure would allow us to fine-tune semantic segmentation systems on less amount of annotated satellite imagery. Furthermore, we show the new state-of-the-art evaluation results of our PP-LinkNet system on three challenging remote sensing datasets.

\bibliographystyle{ACM-Reference-Format}
\bibliography{mylib}

\end{document}